\pdfoutput=1
\documentclass[11pt]{article}

\usepackage{naacl2021}

\usepackage{times}
\usepackage{latexsym}

\usepackage[T1]{fontenc}
\usepackage[utf8]{inputenc}

\usepackage{microtype}

\usepackage{xcolor}
\usepackage{colortbl}
\usepackage{amssymb}
\usepackage{amsmath}
\usepackage{soul}
\usepackage{algorithm}
\usepackage{algorithmic}
\usepackage{booktabs}
\usepackage{graphicx}

\definecolor{green}{RGB}{53, 130, 74}
\definecolor{orange}{RGB}{209, 154, 0}
\definecolor{lightgray}{RGB}{239, 239, 239}
\usepackage{xspace}

\newcommand{\ie}{i.e.\@\xspace}

\newcommand{\textoverline}[1]{$\overline{\mbox{#1}}$}
\newcommand{\errstructure}[1]{\underline{#1}}
\newcommand{\errtarget}[1]{\textoverline{#1}}
\newcommand{\errreversal}[1]{$\overleftrightarrow{\mbox{#1}}$}
\newcommand{\errhallucination}[1]{\st{#1}}

\title{Exploiting Social Media Content for Self-Supervised Style Transfer}

\author{Dana Ruiter$^1$, Thomas Kleinbauer$^1$, Cristina Espa\~{n}a-Bonet$^{2,3}$, \\
{\bf Josef van Genabith$^{2,3}$, Dietrich Klakow$^1$} \\
  $^1$Spoken Language Systems Group, Saarland University, Germany \\
  $^2$Saarland Informatics Campus, Saarland University, Germany \\
  $^3$DFKI GmbH, Germany  \\
  \texttt{druiter@lsv.uni-saarland.de} 
  }

\begin{document}
\maketitle
\begin{abstract}
Recent research on style transfer takes inspiration from unsupervised neural machine translation (UNMT), learning from large amounts of non-parallel data by exploiting cycle consistency loss, back-translation, and denoising autoencoders.
By contrast, the use of self-supervised NMT (SSNMT), which leverages (near) parallel instances hidden in non-parallel data more efficiently than UNMT, has not yet been explored for style transfer. In this paper we present a novel Self-Supervised Style Transfer (3ST) model, which augments SSNMT with UNMT methods in order to identify and efficiently exploit supervisory signals in non-parallel 
social media posts.
We compare 3ST with state-of-the-art (SOTA) style transfer models across civil rephrasing, formality and polarity tasks.
We show that 3ST is able to balance the three major objectives (fluency, content preservation, attribute transfer accuracy) the best, outperforming SOTA models on averaged performance across their tested tasks in automatic and human evaluation.
\end{abstract}

\section{Introduction}

Style transfer is a highly versatile task in natural language processing, where the goal is to modify the stylistic attributes of a text while maintaining its original meaning.
A broad variety of stylistic attributes has been considered, including
formality \citep{rao-tetreault-2018-dear}, gender \citep{prabhumoye-etal-2018-style}, polarity \citep{shen2017style} and civility \citep{laugier-etal-2021-civil}.
Potential industrial applications are manifold and range from simplifying professional language to be intelligible to laypersons \citep{cao-etal-2020-expertise}, the generation of more compelling news headlines \citep{jin-etal-2020-hooks}, to related tasks such as text simplification for children and people with disabilities \citep{martin2020simplification}.

Data-driven style transfer methods can be classified according to the kind of data they use: parallel or non-parallel corpora in the two styles~\citep{jinEtAl:2021}.
To learn style transfer on non-parallel monostylistic corpora, current approaches take inspiration from unsupervised neural machine translation (UNMT) \citep{lample2018unsupervised}, by exploiting cycle consistency loss \citep{lample2018multipleattribute}, iterative back-translation \citep{jin-etal-2019-imat} and denoising autoencoders (DAE) \citep{laugier-etal-2021-civil}.
As these approaches are similar to UNMT they suffer from the same limitations, \ie poor performance relative to supervised neural machine translation (NMT) systems when the amount of UNMT training data is small and/or exhibits domain mismatch \citep{kim-etal-2020-unsupervised}. Unfortunately, this is precisely the case for most existing style transfer corpora.

In this paper, we follow an alternative approach inspired by self-supervised NMT \citep{ruiter-etal-2021-integrating} that jointly learns online (near) parallel sentence pair extraction (SPE), back-translation (BT) and style transfer in a loop. The goal is to identify and exploit supervisory signals present in limited amounts of (possibly domain-mismatched) non-parallel data ignored by UNMT. The architecture of our system--called \textbf{S}elf-\textbf{S}upervised \textbf{S}tyle \textbf{T}ransfer (3ST)--implements an online self-supervisory cycle, where learning SPE enables us to learn style transfer on extracted parallel data, which iteratively improves SPE and BT quality, and thereby style transfer learning, in a virtuous circle.

We evaluate and compare 3ST to current state-of-the-art (SOTA) style transfer models on two established tasks: formality and polarity style transfer, where 3ST is the most balanced model and reaches top overall performance.

To gain insights into the performance of 3ST on an under-explored task, we also focus on the civil rephrasing task, which is interesting as $i$) it has been explored only twice before \citep{nogueira-dos-santos-etal-2018-fighting,laugier-etal-2021-civil} and $ii$) it makes an important societal contribution in order to tackle hateful content online. 
We focus on performance and qualitative analysis of 3ST predictions on this task's test set and identify shortcomings of the currently available data setup for civil rephrasing.
On civil rephrasing, 3ST generates more neutral sentences than the current SOTA model while being on par in overall performance.

Our contribution is threefold: 
\begin{itemize}
    \item  Efficient detection and exploitation of the supervisory signals in non-parallel social media content
    via jointly-learning \textit{online} SPE and BT, 
    outperforming SOTA models on averaged performance across civility, formality and polarity tasks in automatic and human evaluation ($\Delta$ in Tables \ref{t:bleu} and \ref{t:human}).
    \item Simple end-to-end training of a single online model without the need for additional external style-classifiers 
    or external SPE,
    enabling the initialization of the 3ST network on a DAE task, which leads to SOTA-matching fluency scores during human evaluation.
    \item A qualitative analysis that identifies flaws in the current data, emphasizing the need for a high quality civil rephrasing corpus.
\end{itemize}

\section{Related Work}
\label{s:related_work}

\textbf{Style transfer} can be treated as a supervised translation task between two styles \citep{jhamtani-etal-2017-shakespearizing}.
However, for most style transfer tasks, parallel data is scarcely available.
To learn style transfer without parallel data, prior research has focused on exploiting larger amounts of monostylistic data in combination with a smaller amount of style-labeled data. One such approach is using variational autoencoders and disentangled latent spaces \citep{fu2017style}, which can be further incentivized towards generating fluent or style-relevant content by fusing them with adversarial \citep{shen2017style} or style-enforcing \citep{hu2017toward} discriminators. 
\citet{chawla-yang-2020-semi} use a language model as the discriminator, leading to a more informative signal to the generator during training and thus more fluent and stable results.
\citet{li-etal-2018-delete} argue that adversarially learned outputs tend to be low-quality, and that most sentiment modification is based on simple deletion and replacement of relevant words. 

The above approaches focus on separating content and style, either in latent space or surface form, however this separation is difficult to achieve \citep{gonen-goldberg-2019-lipstick}. \citet{dai-etal-2019-style} instead train a transformer together with a discriminator, without disentangling the style features before decoding.
Current approaches treat style transfer similar to an unsupervised neural machine translation \citep{artetxe-etal-2019-effective} task. \citet{jin-etal-2019-imat} create pseudo-parallel corpora by extracting similar sentences offline from two monostylistic corpora to train an initial NMT model which is then iteratively improved using back-translation. \citet{luo2019dual} use a reinforcement approach to further improve sentence fluency. \citet{laugier-etal-2021-civil} improve fluency without the need of any style-specific classifiers, giving their model a head start by initializing it on a pre-trained transformer model. \citet{wang-etal-2020-formality} argue that standard NMT training cannot account for the small differences between informal and formal style transfer, and apply style-specific decoder heads to enforce style differences. 

Our approach differs from the two step approach of \citet{jin-etal-2019-imat}, who first extract similar sentences from style corpora \textit{offline} and then initialize their system by training on them.
\citet{ruiter-etal-2020-self} show that \textit{joint online learning} to extract and translate in self-supervised NMT (SSNMT) leads to higher recall and precision of the extracted data. Following this observation, our 3ST approach
performs similar sentence extraction and style transfer learning online with a single model in a loop.
We further extend the SSNMT-based approach by combining it with UNMT methods, namely by generating additional training data via online back-translation, and by initialising our models with DAE trained in an unsupervised manner.

\section{Self-Supervised Style Transfer (3ST)}
\label{s:method}

\begin{figure}
    \centering
    \includegraphics[width=\columnwidth]{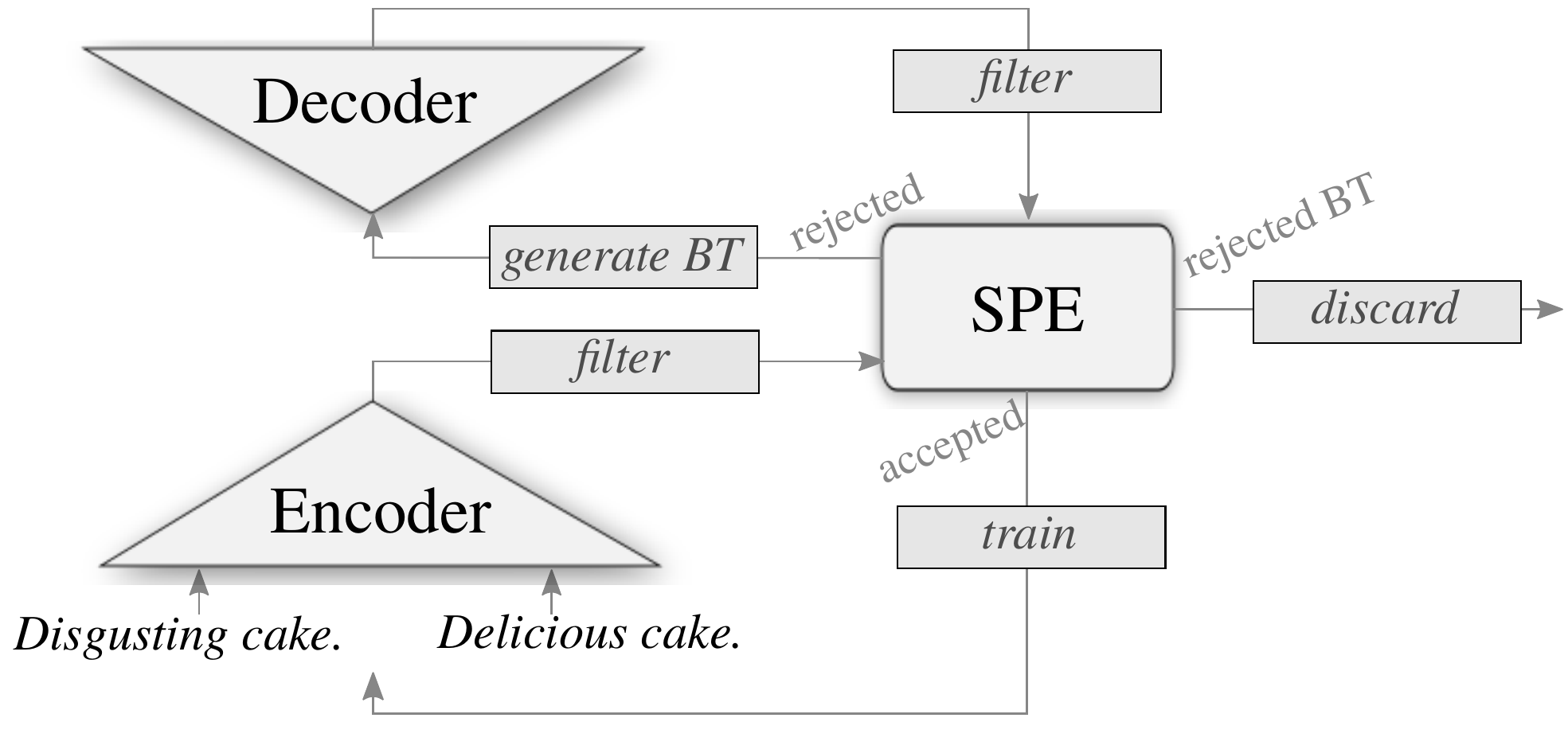}
    \caption{3ST: joint learning of style transfer, SPE, and BT.
    }
    \label{f:architecture}
\end{figure}

Figure \ref{f:architecture} shows the 3ST architecture, which uses the encoder outputs at training time as sentence representations to perform online (near) parallel sentence pair extraction (SPE) together with online back-translation (BT) and style transfer.
\paragraph{Self-Supervised NMT (SSNMT):} SSNMT \citep{ruiter-etal-2019-self} is an encoder-decoder architecture that 
jointly learns to identify parallel data in non-parallel data and bidirectional NMT. 
Instead of using SSNMT on different language corpora to learn machine translation, 
we show how ideas from SSNMT can be used to learn a self-supervised style transfer system from non-parallel social media content.
A single bidirectional encoder simultaneously encodes both styles and maps the internal representations of the two styles into the same space. This way, they can be used to compute similarities between sentence pairs in order to identify similar and discard non-similar ones for training. Formally, given two monostylistic corpora $S1$ and $S2$ of opposing styles, e.g. \emph{toxic} and \emph{neutral}, sentence pairs $(s_{S1} \in S1,s_{S2} \in S2)$
are input to an encoder-decoder system, a transformer in our experiments. 
From the internal representations for the input sentences $s_{S1}$ and $s_{S2}$, SSNMT uses the sum of the word embeddings $w(s)$ and the sum of the encoder outputs $e(s)$ for filtering.  The embedded pairs $\{w(s_{S1}), w(s_{S2})\}$ are scored using the margin-based measure \citep{artetxe2018margin}. The same is done with pairs $\{e(s_{S1}), e(s_{S2})\}$. If a sentence pair is the most similar pair for both style directions \emph{and} for both sentence representations, it is accepted for training, otherwise it is discarded. This sequence of scoring and filtering is denoted as \textbf{sentence pair extraction (SPE)} in 3ST. SPE improves style transfer and style transfer improves SPE online in a virtuous loop, resulting in a single system that jointly learns to identify its supervision signals in the data and to perform style transfer.

\medskip\noindent

To address the characteristics of the monostylistic corpora we extend basic SSNMT in two ways:
\paragraph{Large-Scale Extraction:}
SSNMT extracts parallel data from \textit{comparable corpora}, which contain smaller topic-aligned documents $\{d_{S1},d_{S2}\}$ of similar content, 
thus reducing the search space during SPE from $|S1| \times |S2|$ to $|d_{S1}| \times |d_{S2}|$. However, style transfer corpora usually consist of large collections of (unaligned) sequences of a specific style, which forces the exploration of the full space.
Improving over the one-by-one comparison of vector representations, we index\footnote{%
As our internal representations change during the course of training, we re-index at each iteration over the data.} 
our data using FAISS \citep{faiss2017}.

\paragraph{UNMT-Style Data Augmentation:}
We follow \citet{ruiter-etal-2021-integrating} and use the current models' state to generate \textbf{back-translations} \emph{online} from sentences rejected during SPE in order to increase the amount of supervisory signals to train on. Further, we initialize our style transfer models using \textbf{denoising autoencoding} using BART-style\footnote{This is algorithmically equivalent to using a common pre-trained BART model for initialization, with the benefit that we have full control on the vocabulary size and data it is pre-trained on. We use this benefit by focusing the pre-training on in-domain data instead of generic out-of-domain data.} noise \citep{lewis-etal-2020-bart}. After pre-training a DAE on the stylistic corpora, our models will generate fluent English sentences from the beginning and only need to learn to separate the two styles $S1$ and $S2$ during style transfer learning.

\section{Experimental Setup}
\label{s:experimental_setup}

\subsection{Data}
\label{s:data}

\begin{table}[t]
\small
\centering
\begin{tabular}{l r r r r}
\toprule
\textbf{Corpus} &  \textbf{Train} & \textbf{Dev} & \textbf{Test} & $\varnothing$ \\ 
      \midrule
CivCo-Neutral & 136,618 & 500 & -- & -- \\
CivCo-Toxic & 399,691 & 500 & 4,878 & 14.9 \\
Yahoo-Formal & 1,737,043 & \underline{4,603} & \underline{2,100} & 12.7 \\
Yahoo-Informal & 3,148,351 & \underline{5,665} & \underline{2,741} & 12.4 \\
Yelp-Pos & 266,041 & 2,000 & \underline{500} & 9.9 \\
Yelp-Neg & 177,218 & 2,000 & \underline{500} & 10.7 \\
\bottomrule
\end{tabular}
\caption{Number of sentences of the different tasks train, dev and test splits, as well as average number of tokens per sequence ($\varnothing$) of the tokenized test sets. Splits with target references available are \underline{underlined}. 
}
\label{t:corpora}
\end{table}

\paragraph{Formality}
For the formality task, we use the test and development (dev) splits of the GYAFC corpus \citep{rao-tetreault-2018-dear}, which is based on the Yahoo Answers L6\footnote{\url{www.webscope.sandbox.yahoo.com/catalog.php?datatype=l}} corpus. However, as GYAFC is a parallel corpus and we want to evaluate our models in a setup where only monostylistic data is available, we follow \citet{rao-tetreault-2018-dear} and re-create the training split without downsampling and without creating parallel reference sentences. For this, we extract all answers from the \emph{Entertainment \& Music} and \emph{Family \& Relationships} domains in the Yahoo Answers L6 corpus. We use a BERT classifier fine-tuned on the GYAFC training split to classify sentences as either \emph{informal} or \emph{formal}. This leaves us with a much ($46\times$) larger training split than the parallel GYAFC corpus, although consisting of non-parallel data where a single instance is less informative than a parallel one. We remove sentences from our training data that are matched with a sentence in the official test-dev splits. We deduplicate the test-dev splits to match those used by \citet{jin-etal-2019-imat}. 
For DAE pre-training, we sample sentences from
Yahoo Answers L6.

\paragraph{Polarity}
We use the standard train-dev-test splits\footnote{\url{www.github.com/shentianxiao/language-style-transfer}} of the Yelp sentiment transfer task \citep{shen2017style}. This dataset is already tokenized and lower-cased. Therefore, as opposed to the civility and formality tasks, we do not perform any additional pre-processing on this corpus. 
For DAE pre-training, we sample sentences from a generic Yelp corpus\footnote{\url{www.yelp.com/dataset}} and process them to fit the preprocessing of the Yelp sentiment transfer task, i.e. we lowercase and perform sentence and word tokenization using NLTK  \citep{Loper02nltk:the}.

\paragraph{Civility}
The civil rephrasing task is rooted in the broader domain of hate speech research, which commonly focuses on the detection of hateful, offensive, or profane contents \cite{yang2019xlnet}. 
Besides deletion, moderation, and generating counter-speech \citep{tekiroglu-etal-2020-generating}, which are \emph{reactive} measures after the abuse has already happened, there is a need for \emph{proactive} ways of dealing with hateful contents to prevent harm \citep{jurgens-etal-2019-just}. Civil rephrasing is a novel approach to fight abusive or profane contents by suggesting civil rephrasings to authors before their comments are published. 
So far, civil rephrasing has been explored twice before \citep{nogueira-dos-santos-etal-2018-fighting,laugier-etal-2021-civil}. However, their datasets are not publicly available. In order to compare the works, we reproduce the data sets used in \citet{laugier-etal-2021-civil}.
We follow their approach and create our own train and dev splits on the Civil Comments\footnote{\url{www.tensorflow.org/datasets/catalog/civil_comments}} (CivCo) dataset. 
Style transfer learning requires distinct distributions in the two opposing style corpora. To increase the distinction in our toxic and neutral datasets, we filter them using a list of slurs\footnote{\url{www.cs.cmu.edu/~biglou/resources/bad-words.txt}} such that the toxic portion contains only sentences with at least one slur, and the neutral portion does not contain any slurs in the list. \citet{laugier-etal-2021-civil} kindly provided us with the original test set used in their study. We removed sentences contained in the test set from our corpus and split the remaining sentences into train and dev.
To initialize 3ST on DAE with data related to the civility task domain, i.e. user comments, we sample sentences from generic
Reddit comments crawled with PRAW\footnote{\url{www.praw.readthedocs.io/en/latest/}}.

\paragraph{Preprocessing}

On all datasets, excluding the polarity task data which is already preprocessed, we performed sentence tokenization using NLTK as well as punctuation normalization, tokenization and truecasing using standard Moses scripts
\citep{koehn-etal-2007-moses}.
Following \citet{rao-tetreault-2018-dear}, we remove sentences containing URLs as well as those containing less than $5$ or more than $25$ words. For the civility task only, we allow longer sequences of up to $30$ words due to the higher average sequence length in this task \citep{laugier-etal-2021-civil}. We perform deduplication and language identification using \texttt{polyglot}\footnote{\url{www.github.com/aboSamoor/polyglot}}. We apply a byte-pair encoding \citep{sennrich-etal-2016-neural} 
of $8k$ merge-operations. 
We add target style labels (e.g. \mbox{\emph{\textless pos\textgreater}}) to the beginning of each sequence.
Table \ref{t:corpora} summarizes all train, dev and test splits.

\subsection{Model Specifications}
\label{s:model_specification}
We base our 3ST code on
OpenNMT \citep{klein-etal-2017-opennmt}, using a transformer-base with standard parameters, a batch size of $50$ sentences and a maximum sequence length of $100$ sub-word units.
All models are trained until the attribute transfer accuracy on the development set has converged. Each model is trained on a single Titan X GPU, which takes around $2$--$5$ days for a 3ST model.

For DAE pre-training, we use the task-specific DAE data split into $20M$ train sentences and $5k$ dev and test sentences each. To create the noisy source-side data, we apply BART-style
noise with $\lambda=3.5$ and $p=0.35$ for word sequence masking. We also add one random mask insertion per sequence and perform a sequence permutation.

For BERT classifiers, which we use to automatically evaluate the attribute transfer accuracy, we fine-tune a \texttt{bert-base-cased} model
on the relevant classification task using early stopping with $\delta=0.01$ and patience $5$.

\subsection{Automatic Evaluation}
\label{s:automatic_evaluation_setup}
While 3ST can perform style transfer bidirectionally, we only evaluate on the \emph{toxic}\textrightarrow\emph{neutral} direction of the civility task, as the other direction, i.e. generation of toxic content, would pose a harmful application of our system. Similarly, the formality task is only evaluated for the \emph{informal}\textrightarrow\emph{formal} direction as this is the most common use-case \citep{rao-tetreault-2018-dear}. The polarity task is evaluated in both directions. We compare our model against current SOTA models: multi-class (MUL) and conditional (CON) style transformers by \citet{dai-etal-2019-style}, unsupervised machine translation (UMT) \citep{lample2018multipleattribute}\footnote{Model outputs provided by \citet{he2020a}.} as well as models by \citet{li-etal-2018-delete} (DAR), \citet{jin-etal-2019-imat} (IMT), \citet{laugier-etal-2021-civil} (CAE), \citet{he2020a} (DLA) and \citet{shen2017style} (SCA). 
Our automatic evaluation focuses on four main aspects:

\paragraph{Content Preservation (CP)}
In style transfer, the aim is to change the style of a source sentence into a target style without changing the underlying meaning of the sentence. 
To evaluate CP, BLEU is a common choice, despite its inability to account for paraphrases \citep{wieting-etal-2019-beyond}, which are at the core of style transfer. Instead, we use Siamese Sentence Transformers \footnote{Model \texttt{paraphrase-mpnet-base-v2}} \footnote{\url{https://www.sbert.net/index.html}} to embed the source and prediction and then calculate the cosine similarity.

\paragraph{Attribute Transfer Accuracy (ATA)}
We want to transfer the style of the source sentence to the target style or attributes. Whether this transfer was successful is calculated using a BERT classification model. We train and evaluate our classifiers on the same data splits as the style-transfer models. This yields classifiers with Macro-F1 scores of $93.2$ (formality), $87.4$ (civility) and $97.1$ (polarity) on the task-specific development sets. ATA is the percentage of generated target sentences that were labeled as belonging to the target style by the task-specific classifier. 

\paragraph{Fluency (FLU)}
As generated sentences should be intelligible and natural-sounding to a reader, we take their fluency into consideration during evaluation. The perplexity of a language model is often used to evaluate this \citep{krishna-etal-2020-reformulating}. However, perplexity is unbounded and therefore difficult to interpret, and has the limitation of favoring potentially unnatural sentences containing frequent words \citep{mir-etal-2019-evaluating}. We therefore use a RoBERTa \citep{liu2019roberta} model\footnote{\url{www.huggingface.co/textattack/roberta-base-CoLA}} trained on CoLA \citep{warstadt-etal-2019-neural} to label model predictions as either \emph{grammatical} or \emph{ungrammatical}.

\paragraph{Aggregation (AGG)}

CP, ATA and FLU are important dimensions of style-transfer evaluation. A good style transfer model should be able to perform well across all three metrics. To compare overall style-transfer performance, it is possible to aggregate these metrics into a single value \citep{li-etal-2018-delete}. \citet{krishna-etal-2020-reformulating} show that corpus-level aggregation are less indicative for the overall performance of a system and we thus apply their sentence-level aggregation score, which ensures that each predicted sentence performs well across all measures, while penalizing predictions which are poor in at least one of the metrics. We also report the average AGG difference of a model $m$ to 3ST across all tasks that $m$ was tested on ($\Delta$).

\paragraph{}
The automatic evaluation relies on external models, which are sensitive to hyperparameter choices during training. However, we use the same evaluation models across all style transfer model predictions and supplement the automatic evaluation with a human evaluation. As we observe consistency between the automatic and human evaluation, the underlying models used for the automatic evaluation can be considered to be sufficiently reliable.

\subsection{Human Evaluation}
\label{s:human_evaluation_setup}

We compare the performance of 3ST with each of the two strongest baseline systems per task, 
chosen
based on their aggregated scores
achieved in
the automatic evaluation. 
These are:
CAE and IMT for comparison in the polarity task, DAR and IMT for the formality task and CAE for the civility task. Due to the large number of models in the polarity task, we also include CON and MUL in the human evaluation, as they are strongest on ATA and CP respectively.

For each task, we sample $100$ data points from the original test set and the corresponding predictions of the different models. We randomly duplicate $5$ of the data points to calculate intra-rater agreement, resulting in a total of $105$ evaluation sentences per system. 
Three fluent English speakers
were asked to rate the content preservation, fluency and attribute transfer accuracy of the predictions on a $5$-point Likert scale. In order to aggregate the different values, analogous to the automatic evaluation, we consider the transfer to be \textit{successful} when a prediction was rated with a $4$ or $5$ across all three metrics \citep{li-etal-2018-delete}.
The success rate (SR) is then defined as the ratio of successfully transferred instances over all instances. We also report the cross-task average SR difference of a model to 3ST ($\Delta$).

All inter-rater agreements, calculated using Krippendorff-$\alpha$, lie above $0.7$, except for cases where most samples were annotated repeatedly with the same justified rating (e.g. a continuous FLU rating of $4$) due to the underlying data distribution, which is sanctioned by the Krippendorff measure.
Intra-rater agreement is at an average of $0.928$ across all raters. A more detailed description of the evaluation task and a listing of the task- and rater-specific $\alpha$-values is given in the appendix. 
For the ratings themselves, we calculate pair-wise statistical significance between SOTA models and 3ST using the Wilcoxon T test ($p<0.05$).

\section{Results and Analysis}
\label{s:results_analysis}

\subsection{Automatic Evaluation}
\label{s:automatic_evaluation}

\begin{table}[t]
\small
\centering
\begin{tabular}{l@{\hspace{0.5em}} l r r r >{\columncolor{lightgray}}r r}
\toprule
 \textbf{Task} & \textbf{Model} & \textbf{CP} & \textbf{FLU} & \textbf{ATA} & \textbf{AGG} & \textbf{$\Delta$}\\  \midrule
 Civ. & CAE & *\textbf{64.2} & *\textbf{80.6} & *81.9 & \underline{\textbf{39.8}} & -2.9\\
 & 3ST & 60.5 & 75.3 & \textbf{89.7} & \textbf{39.0} & \textbf{0.0}  \\ \midrule
For. & DAR & *64.5 & *27.9 & *66.0 & *\underline{14.2} & -30.0\\
 & IMT & *71.5 & *73.1 & *79.2 & *\underline{45.2} & -7.6 \\
 & SCA & *54.4 & *14.7 & *27.4 & *4.0 & -40.3 \\
 & 3ST & \textbf{75.6} & \textbf{83.1} & \textbf{84.9} & \textbf{54.7} & \textbf{0.0} \\ \midrule
 Pol. & CAE & *48.3 & *76.4 & *84.3 & *\underline{28.7} & -2.9 \\
 & CON & *57.5 & *32.5 & *\underline{\textbf{91.3}} & *17.3 & -18.0 \\
 & DAR & *50.4 & *32.7 & *87.8 & *15.8 & -30.0 \\
 & DLS & *50.9 & *50.4 & 85.3 & *20.1 & -15.2 \\
 & IMT & *42.5 & *\underline{\textbf{84.4}} & *84.6 & *\underline{29.6}  & -7.6 \\
 & MUL & *\underline{\textbf{62.6}} & *42.3 & *82.5 & *20.4 & -14.9 \\
 & SCA & *36.7 & *19.5 & *73.2 & *5.5 & -40.3 \\
 & UMT & *54.8 & *55.7 & 85.4 & *24.2 & -11.1 \\
 & 3ST & 55.7 & 81.0 &  85.4 & \textbf{35.3} & \textbf{0.0} \\
\bottomrule
\end{tabular}
\caption{Automatic scores for CP, FLU, ATA and their aggregated score (AGG) of SOTA models and our approach (3ST) across the Civ(ility), For(mality) and Pol(arity) tasks. Cross-task average AGG difference to 3ST under $\Delta$.
Best values per task in \textbf{bold} and models selected for human evaluation \underline{underlined}. Values statistically significantly different ($p<0.05$) from 3ST are marked with *.}
\label{t:bleu}
\end{table}

Table~\ref{t:bleu} provides an overview of the CP, FLU, ATA and AGG results of all compared models across the three tasks.

\paragraph{Civility}
On attribute transfer accuracy, 3ST improves by $+7.8$ points over CAE, while CAE is stronger in content preservation ($+3.7$) and fluency ($+5.3$). There is, however, no statistically significant difference in the overall aggregated performance of the models, indicating that they are equivalent in performance.

\paragraph{Formality}
3ST substantially outperforms SOTA models in all four categories, with an overall performance (AGG) that surpasses the top-scoring SOTA model (IMT) by $+9.5$ points. This is indicative, as IMT was trained on a shuffled version of the parallel GYAFC corpus, which contains highly informative human written paraphrases, 
while 3ST was trained on a truly non-parallel corpus.

\paragraph{Polarity}
The polarity task has more recent SOTAs to compare to, and the results show that no single model is best in all three categories. While MUL is strongest in content preservation ($62.6$), its fluency is low and outperformed by 3ST by $+38.7$ points, leading to a much lower overall performance (AGG) in comparison to 3ST ($+14.9$). Similarly, CON is strongest in attribute transfer accuracy ($91.3$) but has a low fluency ($32.5$), leading to a lower aggregated score than 3ST ($+18$). IMT is the strongest SOTA model with an overall performance (AGG) of $29.6$ and the highest fluency score ($84.4$). Nevertheless, it is outperformed by 3ST by $+5.7$ points on overall performance (AGG), which is due to the comparatively better performance in content preservation ($+13.2$) of 3ST. 
Interestingly, unsupervised NMT (UMT) performs equally well on attribute transfer accuracy, while being slightly outperformed by 3ST in content preservation ($+0.9$). This may be due to the information-rich parallel instances automatically found in training by the SPE module. Further, 3ST has a much higher fluency than UMT ($+25.3$), which is due to its DAE pre-training.
While 3ST is not top-performing in any of the three metrics CP, FLU and ATA, its top-scoring overall performance (AGG) shows that it is the most balanced model.

\paragraph{Overall Trends}
Table \ref{t:bleu} shows that 3ST outperforms each of the SOTA models fielded in a single task (CON, DLS, MUL, UMT) by the respective AGG $\Delta$, and all other models (CAE, DAR, IMT, SCA) on average AGG $\Delta$\footnote{e.g. $\Delta({\rm DAR,3ST}) = \frac{14.2+15.8}{2} - \frac{54.7+35.3}{2} = -30$ across Formality and Polarity.}. 
3ST achieves high levels of FLU, with ATA in the medium to high 80’s, clear testimony to successful style transfer.

\subsection{Human Evaluation}
\label{s:human_evaluation}
\begin{table}[t]
\small
\centering
\begin{tabular}{l@{\hspace{0.8em}} l r r r >{\columncolor{lightgray}}r r}
\toprule
\textbf{Task} & \textbf{Model} & \textbf{CP} & \textbf{FLU} & \textbf{ATA} & \textbf{SR} & \textbf{$\Delta$} \\ 
      \midrule
Civ. & CAE & \textbf{2.97} & 4.01 & *2.50 & 17.0 & -8.5 \\
& 3ST & 2.80 & \textbf{4.05} & \textbf{3.03} & \textbf{21.0} & \textbf{0.0} \\ \midrule
For. & DAR & *2.75 & *2.87 & 2.72 & 3.0 & -8.0 \\
& IMT & 3.49 & 4.10 & \textbf{2.83} & 5.0 & -13.0 \\
& 3ST & \textbf{3.75} & \textbf{4.29} & 2.82 & \textbf{11.0} & \textbf{0.0} \\ \midrule
Pol. & CAE & *3.64 & 4.46 & 3.90 & 54.0 & -8.5 \\
& CON & 4.20 & *3.47 & 3.97 & 44.0 & -23.0 \\
& IMT & *3.54 & \textbf{4.68} & 3.84 & 47.0 & -13.0 \\
& MUL & *\textbf{4.34} & *3.66 & 3.68 & 41.0 & -26.0 \\
& 3ST & 3.99 & 4.58 & \textbf{4.03} & \textbf{67.0} & \textbf{0.0}  \\
\bottomrule
\end{tabular}
\caption{Average human ratings of CP, FLU, ATA and success rate (SR) on the three transfer tasks Civ(ility), For(mality) and Pol(arity). Cross-task average SR difference to 3ST ($\Delta$). Best values per task in \textbf{bold}. Values statistically significantly different ($p<0.05$) from 3ST are marked with *.}
\label{t:human}
\end{table}

Human evaluation shows that 3ST has a high level of \textbf{fluency}, as it either outperforms or is on par with current SOTA models across all three tasks (Table \ref{t:human}),
with ratings between $4.05$ (civility) and $4.58$ (polarity), and gains of up $+1.42$ (DAR, formality) points. According to the annotation protocol, a rating of 4 and 5 is to describe content written by native speakers, thus annotators deemed most generated sentences to have been written by a native speaker of English. 

For \textbf{content preservation} and \textbf{attribute transfer}, there seems to be a trade-off. In the formality task, 3ST outperforms or is on par with current SOTAs on CP with gains between $+0.26$ (IMT) and  $+1.0$ (DAR) points, and ATA is on par with the SOTA ($-0.01$, IMT). Note that for all models tested on the formality task, the success rate is low. This is due to the nature of the training data, where many sentences in the formal portion of the dataset tend to be rather neutral, i.e. neither formal nor informal, rather than truly formal sentences. For the civility task, on the other hand, 3ST outperforms the current SOTA on ATA with gains of $+0.53$ (CAE) while being on par on CP ($-0.17$). 
For the polarity task, the CP is slightly below the best model ($-0.35$, MUL).

While some models are strong on single values, 3ST has the highest success rate (SR) across all tasks. 3ST outperforms each of the single task models (DAR, CON, MUL) on SR by $\Delta$ and each of the multitask models (CAE, IMT) by average cross-task SR $\Delta$, again highlighting that it balances best between the three capabilities CP, FLU and ATA, which leads to best-performing style transfer predictions.

\subsection{Qualitative Analysis}

\begin{table}[t]
\small
\centering
\begin{tabular}{l@{\hspace{0.17cm}}l@{\hspace{0.05cm}}l}
\toprule
&SRC & \textit{What our ignorant PM, Mad McCallum and} \\ & & \textit{stupid Liberal politicians going to say?} \\
(1) &CAE & \textit{what our pm, \errhallucination{trudeau} and his liberals are going} \\ & & \textit{to do about this?\errhallucination{... ... .. .. .. .. .. .. .. .. .}} \\
&3ST & \textit{\errstructure{Mad McCallum, what are our politicians}} \\ & & \textit{\errstructure{going to say?}} \\ \midrule
&SRC & \textit{Dear Hipster Jackass- Go to Bend.} \\
(2) &CAE & \textit{dear hippie \errhallucination{- go to hawaiian to get around........}} \\
&3ST & \textit{Dear Hipster \errtarget{Jackass}- Go to Bend.} \\ \midrule
&SRC & \textit{Trump's a liar.} \\
(3) &CAE & \textit{trump's a \errtarget{liar}.} \\
&3ST & \textit{Trump's a \errreversal{good man}.} \\ \midrule
&SRC & \textit{Says the idiot on perpetual welfare.}\\
(4) &CAE & \textit{says the \errhallucination{author on the daily basis, on the basis}} \\ & & \textit{of perpetual welfare.} \\
&3ST & \textit{\textbf{Says the guy on perpetual welfare.}}\\ \midrule
&SRC & \textit{A muslim racist.} \\
(5) &CAE & \textit{a muslim \errreversal{minority}.} \\
&3ST & \errhallucination{Not a democrat.} \\ \midrule
&SRC & \textit{Quit trying to justify what this jackass did.} \\
(6) &CAE & \textit{quit trying to justify what this \errtarget{jackass} did.} \\
&3ST & \textit{\textbf{Quit trying to justify what he did.}} \\ \midrule
&SRC & \textit{There was no consensus, 1 idiot and everyone} \\ & &  \textit{else in the situation let him know he was in} \\ & & \textit{the wrong.} \\
(7) &CAE & \textit{there was no consensus, no one in the room} \\ & & \textit{\errstructure{and everyone in the room} knew he was in} \\ & & \textit{the wrong place.} \\
&3ST & \textit{\textbf{No, there was no consensus in the past, and}} \\ & & \textit{\textbf{everyone else knew he was in the wrong place.}} \\ 
\bottomrule
\end{tabular}
\caption{3ST and SOTA model (CAE) predictions on the CivCo test set, with \textbf{adequate} predictions, error in \errstructure{structure}, \errtarget{target attribute}, \errreversal{stance reversal}, and \errhallucination{hallucinations} marked.}
\label{t:qualitative}
\end{table}

For our qualitative analysis, we focus on the civility task as this is a challenging, novel task and we want to understand its limitations. We analyze the same subset of the test set used for human evaluation and annotate common mistakes. Common errors in the neutral counterparts generated by 3ST can be classified into four classes. We observe fluency or \emph{structural errors} ($11\%$ of sentences), e.g. a subject becoming a direct form of address (Table \ref{t:qualitative}, Ex-1). \emph{Attribute errors} ($14\%$) (Ex-2), where toxic content was not successfully removed, are another common source of error. Similarly to \citet{laugier-etal-2021-civil}, we observe \emph{stance reversal} ($14\%$), i.e. where a usually negative opinion in the original source sentence is reversed to a positive polarity (Ex-3). This is due to a negativity bias on the toxic side of the CivCo corpus, while the neutral side contains more positive sentences, thus introducing an incentive to translate negative sentiment to positive sentiment. Unlike \citet{laugier-etal-2021-civil}, we do not observe that hallucinations are most frequent at the end of a sequence (\emph{supererogation}).
Rather, \emph{related hallucinations},
where unnecessary content is mixed with words from the original source sentence, are found at arbitrary positions ($23\%$, Ex-4, CAE).
We observe few hallucinations where a prediction has no relation with the source ($4\%$, Ex-5).

Phenomena such as hallucinations can become amplified through back-translation \citep{raunak2021curious}. However, as they are most prevalent in the civility task,
hallucinations in this case are likely originally triggered by long source sentences that $i)$ overwhelm the current models' capacity, and $ii)$ add additional noise to the training. It is less likely that a complex sentence has a perfect rephrasing to match with and therefore instead it will match with a similar rephrasing that introduces additional content, i.e.\@ noise. For reference, the average length of source sentences that triggered hallucinations was $21.9$ words, while for adequate re-writings ($39\%$), it was $8$ words. Note that we capped sentence lengths to $30$ words in the training data
while the test data contained sentences with up to $85$ words.

Successful rephrasings 
are usually due to one of two factors.
3ST either
\textit{replaces} profane words by their neutral counterparts (Ex-\{4,6\}) or \textit{removes} them (Ex-7).

\subsection{Ablation Study}
\label{s:ablation}

\begin{table}[t]
\small
\centering
\begin{tabular}{l l r r r >{\columncolor{lightgray}}r}
\toprule
 \textbf{Task} & \textbf{Model} & \textbf{CP} & \textbf{FLU} & \textbf{ATA} & \textbf{AGG} \\ \midrule
 Civ. & 3ST     & 60.5 & \textbf{75.3} & 89.7 & \textbf{39.0} \\ 
 & -SPE    & *\textbf{89.5} & *39.4 & *12.1 & *3.7 \\
 & -BT     & *44.4 & *59.4 & 90.3 & *22.8 \\
 & -DAE    & *36.8 & *43.3 & *\textbf{97.5} & *15.7 \\
 & -BT-DAE & *37.8 & *43.8 & *95.3 & *16.4 \\ \midrule
 For. & 3ST     & 75.6 & 83.1 & 84.9 & \textbf{54.7} \\
 & -SPE & *\textbf{99.3} & *73.4 & *17.7 & *14.8 \\
 & -BT    & *66.4 & *\textbf{85.1} & *92.6 & *52.8 \\
 & -DAE   & *55.7 & *64.2 & *93.1 & *35.1 \\
 & -BT-DAE & *57.8 & *79.5 & \textbf{*94.0} & *44.5 \\ \midrule
 Pol. & 3ST & 55.7 & \textbf{81.0} & 85.4 & \textbf{35.3} \\
 & -SPE & *\textbf{100.0} & *80.5 & *2.9 & *1.9 \\
 & -BT & *44.0 & *79.0 & *88.3 & *29.2 \\
 & -DAE & *29.8 & *43.6 & *89.7 & *11.6 \\ 
 & -BT-DAE & *38.0 & *63.3 & *\textbf{91.1} & *21.5 \\
\bottomrule
\end{tabular}
\caption{3ST Ablation. CP, FLU and ATA with SPE, BT, DAE removed. Best values per task in \textbf{bold}.
}
\label{t:bleu_ablation}
\end{table}

To analyze the contribution of the three main components (SPE, BT and DAE) of 3ST, we remove them individually from the original architecture and observe the performance of the resulting models on the three different tasks (Table \ref{t:bleu_ablation}).
Without SPE, the model merely copies source sentences without performing style transfer, resulting in a large drop in overall performance (AGG).
This shows in the low ATA scores ($1.9$--$14.8$), which are in direct correlation with the extremely high scores in CP ($89.5$--$100.0$) achieved by this model. 
This underlines that SPE is vital to the style-transfer capabilities of 3ST, as it retrieves similar paraphrases from the style corpora and lets 3ST train on these. This pushes the system to generate back-translations which themselves are paraphrases that fulfill the style-transfer task.
BT and DAE are integral parts of 3ST, too, that improve over the underlying self-supervised neural machine translation (-BT-DAE) approach. 
This can be seen in the drastic drops of CP and FLU scores
when BT and DAE techniques are removed. Especially DAE is important for the fluency of the model. The gains in CP and FLU through BT and DAE come at a minor drop in ATA.

\section{Conclusion}
\label{s:conclusion}

3ST is a style transfer architecture that efficiently uses the supervisory signals present in non-parallel social media content,
by $i)$ jointly learning style transfer and similar sentence extraction during training, $ii)$ using online back-translation and $iii)$ DAE-based initialization. 
3ST gains strong results on all three metrics FLU, ATA and CP, 
outperforming SOTA models on averaged performance ($\Delta$) across their tested tasks in automatic (AGG) and human (SR) evaluation.
We present one of the first studies on automatic civil rephrasing 
and, importantly, identify current weaknesses in the data,
which lead to limitations in 3ST and other SOTA models on the civil rephrasing task.
Our code and model predictions are publicly available at \url{https://github.com/uds-lsv/3ST}.

\section*{Acknowledgements}
We want to thank the annotators for their keen work. 
Partially funded by the DFG (WI 4204/3-1), German Federal Ministry of Education and Research (01IW20010) and the EU Horizon 2020 project \href{http://www.compriseh2020.eu/}{COMPRISE} (3081705).
The author is responsible for the content of this publication.

\bibliography{naacl2021}
\bibliographystyle{acl_natbib}

\appendix

\section{Human Evaluation Task}
\label{s:rater_agreement}

We perform a human evaluation to assess the quality of the top performing models according to automatic metrics.

We select 3 systems for Formality, 5 systems for Polarity and the only 2 systems available for the Civility task.
For each of these tasks, we sample $100$ data points from the original test set and the corresponding predictions of the different models. We randomly duplicate 5 of the points for quality controls, resulting in evaluation tests with 105 sentences per system. 
Three fluent English speakers (\emph{raters}) were shown with pairs source--system prediction and were asked to rate the content preservation, fluency and attribute transfer accuracy of the predictions on a $5$-point Likert scale. Raters were payed around 10 Euros per hour of work.

We calculate the reliability of the ratings using Krippendorff-$\alpha$~\cite{Krippendorff:04}. Table~\ref{t:alphas} shows the inter-rater agreement measured by $\alpha$ for content preservation (CP), fluency (FLU) and attribute transfer accuracy (ATA).
Notice that $\alpha$ significantly differs between tasks. The lower $\alpha$ on polarity CP and formality task ATA is due to the repetitive ratings of the same kind. i.e. $4,5$ on polarity CP and $3$ for formality ATA, which is sanctioned by the Krippendorff measure. 
For the intra-rater agreement estimated from 40 duplicated sentences per rater, we obtain  values of 0.988 (Rater-1),  0.869 (Rater-2) and 0.927 (Rater-3).

\begin{table}[t!]
\small
\centering
\begin{tabular}{l r r r}
\toprule
 \textbf{Task} & \multicolumn{3}{c}{\textbf{Krippendorff-$\mathbf{\alpha}$}}\\ 
 & CP & FLU & ATA \\\midrule
Civility & 0.744 & 0.579 & 0.688 \\
Formality & 0.751 & 0.718 & 0.352 \\
Polarity & 0.426 & 0.705 & 0.837 \\
\bottomrule
\end{tabular}
\caption{Inter-rater agreement calculated using Krippendorff-$\alpha$ across the different tasks and metrics.
}
\label{t:alphas}
\end{table}

\section{Performance Evolution}

\begin{figure}
    \centering
    \includegraphics[width=\columnwidth]{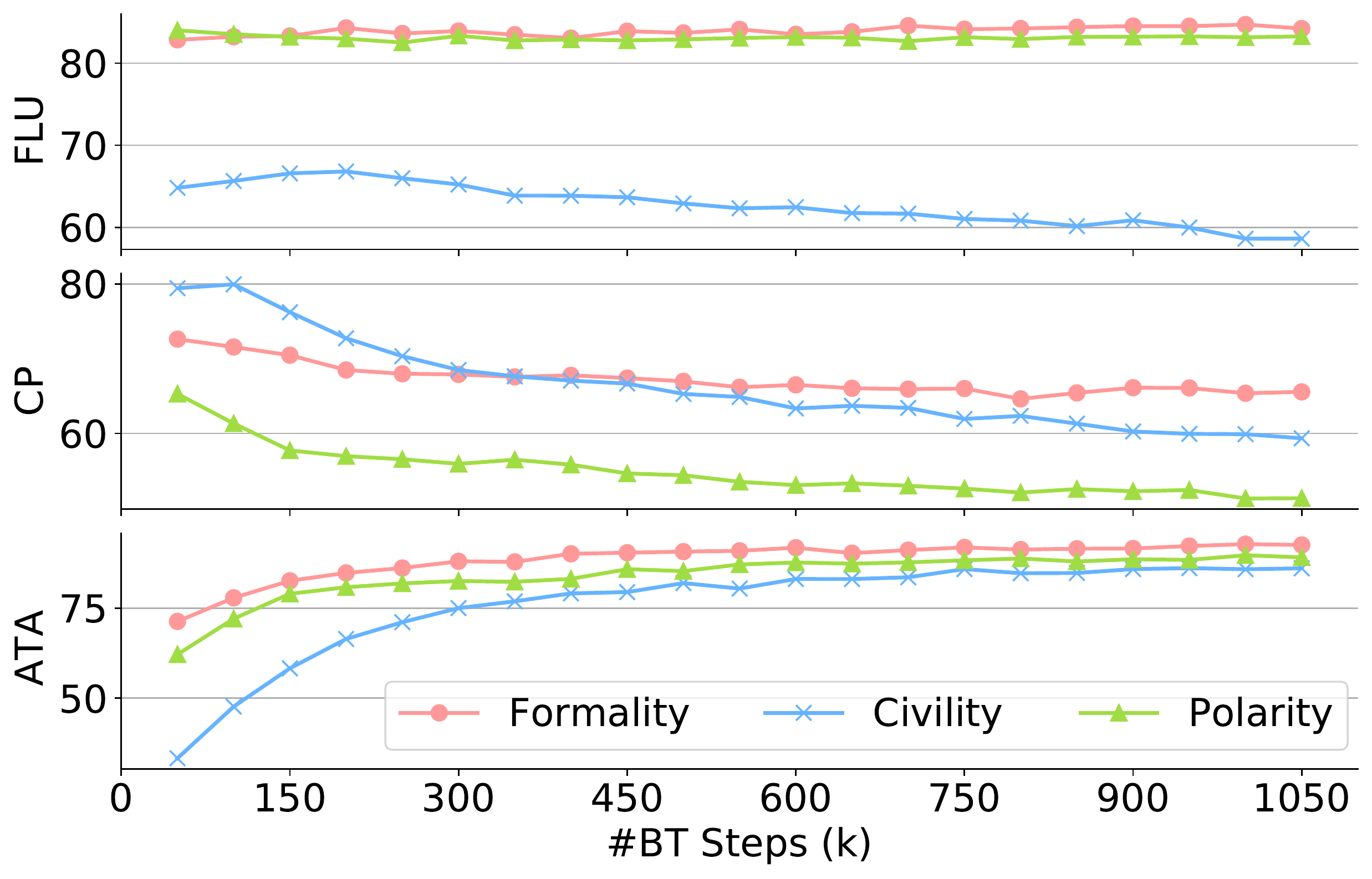}
    \caption{FLU, CP and ATA of generated back-translations (BTs) during training of 3ST on the three transfer tasks.}
    \label{f:bt_steps}
\end{figure}

The back-translations that 3ST generates during training give us a direct insight into the changing state of the model throughout the training process. We thus automatically evaluate ATA, FLU and CP on the back-translations over time.

BT \textbf{fluency} (Figure \ref{f:bt_steps}, top) on all three tasks is strong already at the beginning of training, due to the DAE pre-training. For the formality and polarity task, the high level of FLU 
remains stable
($\sim 80$) throughout training, while for Civility it slightly drops.
This underlines the observation that the Civility task is prone to hallucinations due to the sparse amount of parallel supervisory signals in the dataset, which then leads to lower FLU scores.

For all tasks, \textbf{content preservation} 
between the generated BTs and the source sentences is already high at the beginning of training. This is due to the DAE pre-training which taught the models to copy and denoise inputs.
All of the models decay in CP over time,
showing that they are slowly diverging from merely copying 
inputs. CP scores of the formality and the polarity tasks are close to convergence at around $1M$ train steps, while the scores of the civility task keep on decaying. This may again be due to the complexity of the data of the toxicity task, which contains longer sequences than the other two. This can lead to hallucinations when supervisory signals are lacking.

As back-translation CP decays, \textbf{attribute transfer} accuracy increases dramatically. Especially on the civility task, where the initial accuracy is low ($8.2\%$) but grows to ATA ${\sim}82\%$.
For the other two tasks, the curves are less steep, and most of the transfer is learned at the beginning, within the first $300k$ generated BTs, after which they converge with ATA ${\sim}95\%$ (formality) and ${\sim}88\%$ (polarity). This shows the trade-off between attribute accuracy and content preservation: the higher the ATA, the lower the CP score. Nevertheless, as ATA converges earlier than CP (for formality and polarity tasks), an earlier training stop can easily benefit content preservation while having little impact on the already converged ATA.

\section{Sample Predictions}
\label{s:appendix_predictions}

\begin{table*}[t]
\small
\centering
\begin{tabular}{l@{\hspace{0.05cm}}l}
\toprule
\multicolumn{2}{l}{\emph{Civility}} \\
SRC & \textit{It is time to impeach this idiot judge.} \\
CAE & \textit{it is time to impeach this judge.} \\
3ST & \textit{It is time to impeach this judge.} \\ \midrule
SRC & \textit{This is classic example of collective corporate stupidity and individual managerial malice.} \\
CAE & \textit{this is classic case of corporate welfare and collective bargaining.} \\
3ST & \textit{This is classic example of collective corporate greed and individual managerial malice.} \\ \midrule
SRC & \textit{You silly goose!} \\
CAE & \textit{you mean the goose, right?} \\
3ST & \textit{You forgot the goose!} \\ \midrule
SRC & \textit{Afraid of how idiotic social engineering makes people look?} \\
CAE & \textit{imagine how socially acceptable some of the people make?} \\
3ST & \textit{Afraid of how social engineering works.} \\ \midrule
SRC & \textit{Stupid idea.} \\
CAE & \textit{no idea..............} \\
3ST & \textit{Not a good idea.} \\ \midrule \midrule
\multicolumn{2}{l}{\emph{Formality}} \\
SRC & \textit{haha julesac is funny, but mean.} \\
DAR & \textit{is funny , but I understand what you mean .} \\
IMT & \textit{That is funny . Those silly people annoy me !} \\
3ST & \textit{Julesac is very funny.} \\ \midrule
SRC & \textit{DON'T LET HER RULE YOUR LIFE, SHE WILL JUST HAVE TO LEARN TO DEAL WITH IT.} \\
DAR & \textit{LET HER BE , SHE WILL LEARN TO DEAL WITH IT .} \\
IMT & \textit{TELL HER YOUR TRUE FEELINGS , IT MAY SHOCK HER BUT WILL WORK .} \\
3ST & \textit{Do NOT LET HER RUN WITH YOU, SHE WILL NEVER HAVE TO WORK.} \\ \midrule
SRC & \textit{cause it's buy one take one.} \\
DAR & \textit{I can not wait to buy one take one .} \\
IMT & \textit{Because it is buy one take one . } \\
3ST & \textit{You can buy one.} \\ \midrule
SRC & \textit{All my votes are going to Taylor Hicks though...} \\
DAR & \textit{All my votes are , and I am going to Hicks} \\
IMT & \textit{All my votes are going to Taylor .} \\
3ST & \textit{All my votes are going to be Taylor Hicks.} \\ \midrule
SRC & \textit{but paris hilton isn't far behind.} \\
DAR & \textit{I do not know but is n't far behind .} \\
IMT & \textit{I ca n't read the stars , just find another way to say it .} \\
3ST & \textit{Paris hilton is far behind.} \\ \midrule \midrule
\multicolumn{2}{l}{\emph{Polarity}} \\
SRC & \textit{even if i was insanely drunk , i could n't force this pizza down .} \\
CAE & \textit{even if i was n't in the mood , i loved this place . } \\
IMT & \textit{honestly , i could n't stop eating it because it was so good !} \\
3ST & \textit{even if i was drunk , i could still force myself .} \\ \midrule
SRC & \textit{i will definitely return often !} \\
CAE & \textit{i will not return often ! ! ! ! } \\
IMT & \textit{i will definitely not return !} \\
3ST & \textit{i will not return often !} \\ \midrule
SRC & \textit{no massage with my manicure or pedicure .} \\
CAE & \textit{great massage with great pedicure and manicure .} \\
IMT & \textit{awesome relaxation and massage with my pedicure .} \\
3ST & \textit{great massage with my manicure and pedicure .} \\ \midrule
SRC & \textit{excellent knowledgeable dentist and staff !} \\
CAE & \textit{excellent dentist and dental hygienist ! ! ! !} \\
IMT & \textit{not very knowledgeable staff !} \\
3ST & \textit{horrible dentist and staff !} \\ \midrule
SRC & \textit{do not go here if you are interested in eating good food .} \\
CAE & \textit{definitely recommend this place if you are looking for good food at a good price .} \\
IMT & \textit{if you are looking for consistent delicious food go here .} \\
3ST & \textit{if you are looking for good food , this is the place to go .} \\
\bottomrule
\end{tabular}
\caption{Examples of 3ST and SOTA model predictions.}
\label{t:appendix_predictions}
\end{table*}

For each of the three tasks, Civility, Formality and Polarity, we randomly sample 5 source sentences from the respective test sets. In Table \ref{t:appendix_predictions} we present these source sentences together with the corresponding prediction of 3ST and the two best-scoring SOTA models with respect to the AGG score per task, namely CAE for Civility, DAR and IMT for Formality and CAE and IMT for Polarity.

\end{document}